# LEARNING-BASED PLANNING FOR IMPROVING SCIENCE RETURN OF EARTH OBSERVATION SATELLITES

[1]*Abigail Breitfeld, [2]Alberto Candela, [2]Juan Delfa, [3]Akseli Kangaslahti, [2]Itai Zilberstein, [2]Steve Chien, [1]David Wettergreen

*lead presenter

[1]abreitfe@andrew.cmu.edu, Carnegie Mellon University, USA

[2]Jet Propulsion Laboratory, California Institute of Technology, USA

[3]University of Michigan, USA

***Abstract***–Earth observing satellites are powerful tools for collecting scientific information about our planet, however they have limitations: they cannot easily deviate from their orbital trajectories, their sensors have a limited field of view, and pointing and operating these sensors can take a large amount of the spacecraft's resources. It is important for these satellites to optimize the data they collect and include only the most important or informative measurements. Dynamic targeting is an emerging concept in which satellite resources and data from a lookahead instrument are used to intelligently reconfigure and point a primary instrument. Simulation studies have shown that dynamic targeting increases the amount of scientific information gathered versus conventional sampling strategies. In this work, we present two different learning-based approaches to dynamic targeting, using reinforcement and imitation learning, respectively. These learning methods build on a dynamic programming solution to plan a sequence of sampling locations. We evaluate our approaches against existing heuristic methods for dynamic targeting, showing the benefits of using learning for this application. Imitation learning performs on average 10.0% better than the best heuristic method, while reinforcement learning performs on average 13.7% better. We also show that both learning methods can be trained effectively with small amounts of data.

# 1 INTRODUCTION

Earth observing satellites serve an important purpose in acquiring scientific data, with applications in various fields such as geology, meteorology, and climate science. However, these satellites have limitations in terms of data collection. They are confined to predetermined orbital paths, possess sensors with restricted fields of view, and necessitate a significant portion of the spacecraft's energy for sensor operation and positioning. Therefore, it is imperative for these satellites to optimize their data collection by focusing on the most informative measurements.

The primary objective of this work is to develop strategies for determining the most efficient way for Earth observation satellites to sample their surroundings while considering practical constraints, particularly power consumption. We concentrate on the concept of dynamic targeting, i.e. developing intelligent methods for orienting satellite instruments to enhance their scientific output. Prior research has indicated that employing dynamic targeting methods results in a higher volume of informative scientific data compared to uniform or random sampling techniques. Previous approaches to this challenge have used various methodologies such as greedy algorithms, local search techniques, constraint programming, and dynamic programming (Candela et al. 2023). In this work, we introduce a novel approach to dynamic targeting that leverages two learning frameworks to address this problem, reinforcement and imitation learning.

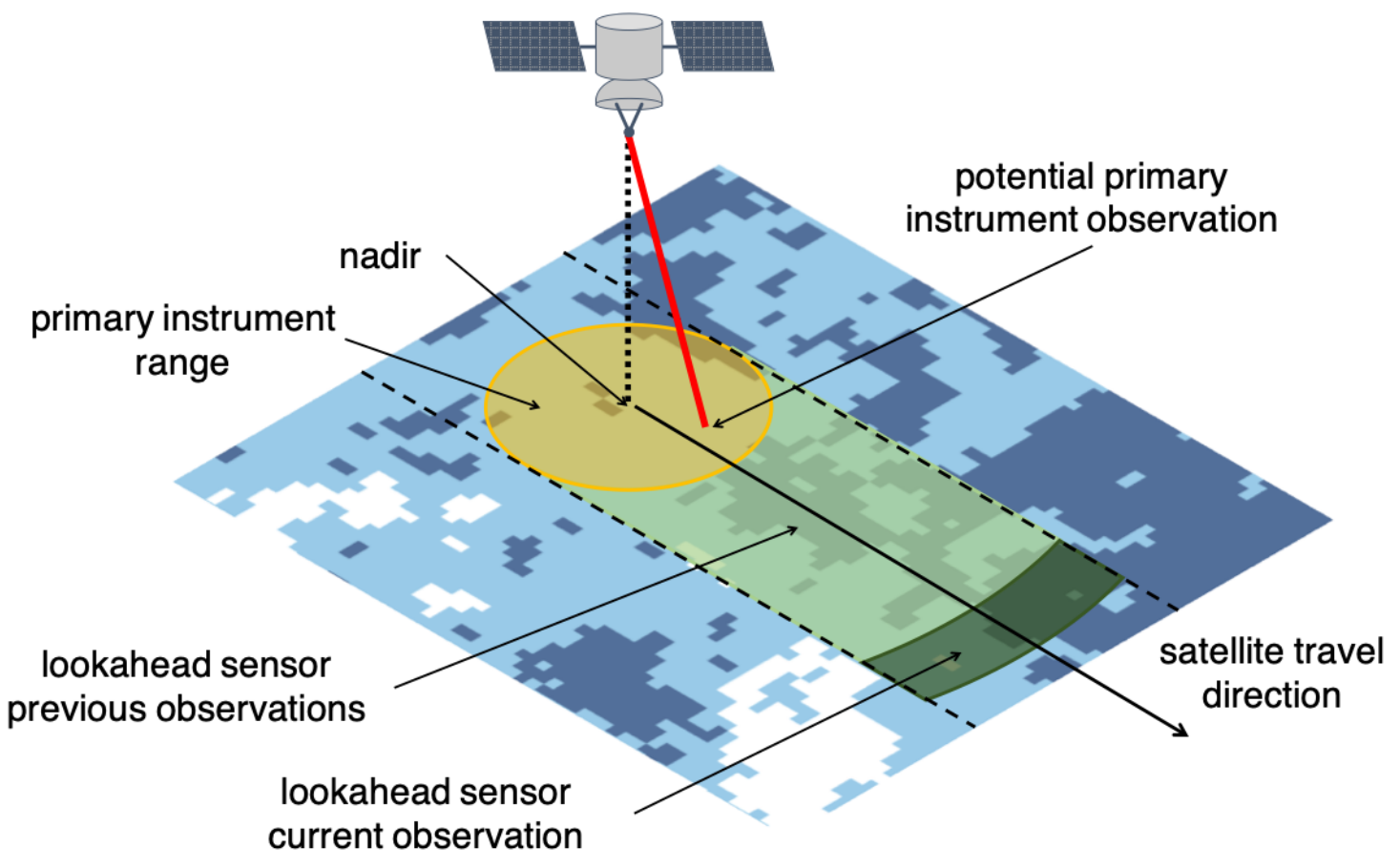

*Figure 1: An Earth observation satellite with a primary sensor and a lookahead sensor that gives information about the environment that lies on the path ahead. The goal of this work is to determine where to point the primary sensor to measure the highest value scientific targets (Candela et al. 2023).*

# 2 BACKGROUND

## 2.1 Related Work

Satellites equipped with agile (pointable) instruments offer a more efficient approach to data collection compared to those featuring static "pushbroom" sensors. The Eagle Eye project has developed planning and scheduling models for observations of pointable 2D satellite instruments (Knight, Donnellan, and Green 2013), and Lemaître et al. focus on scheduling for Agile Earth Observing Satellites (AEOS) which have three degrees of freedom for image acquisition (Lemaître et al. 2002). However, selecting observations for highly-agile instruments is an ongoing challenge; various techniques have been proposed to address this problem, including greedy algorithms, constraint programming approaches, dynamic programming, and local search methods. It is important to note that existing works on agile instruments generally do not leverage past observations.

Liao and Yang present a strategy for scheduling the order of imaging operations for FormoSat-2, a low-earth-orbit remote sensing satellite that previously imaged Taiwan (Liao and Yang 2005). This approach considers current and upcoming weather conditions to provide a plan, which is periodically rescheduled using a rolling horizon scheme.

The Autonomous Sciencecraft Experiment (ASE) was used to analyze acquired imagery and schedule future observations on the Earth Observing One spacecraft for over a decade (Chien et al. 2005). ASE operated on the orbital timescale (roughly 90 minutes), whereas ours and other methods can respond within minutes or faster (Candela et al. 2023). A prototype of a heuristic-based spacecraft pointing scheduler is presented by Chien and Troesch, which operates on the order of one overflight (about 8 minutes) (Chien and Troesch 2015).

The German FireBIRD autonomy mission concept involves two satellites (TET-1 and BIROS), which monitor fires from space with high accuracy. FireBIRD is devised to enhance existing low-resolution systems (up to 1 km/pixel) such as MODIS, METEOSAT, CALIPSO, ADM, and EarthCARE. Lenzen et al. propose a planning and scheduling autonomy concept within the FireBIRD mission scope that can immediately react to onboard-detected events, which combines the advantages of onboard and on-ground scheduling (Lenzen et al. 2014). However, this strategy has not yet flown onboard a spacecraft.

Beaumet, Verfaillie, and Charmeau conducted a study analyzing the feasibility of using an on-board lookahead camera to detect clouds in combination with a primary instrument (Beaumet,

Verfaillie, and Charmeau 2011). Their work proposes an online algorithm that combines both instruments for autonomous high-level mission management, which increases satisfaction of mission objectives.

Rapid cloud screening has been used to reduce data volume on the Airborne Visible-Infrared Imaging Spectrometer (AVIRIS-NG) instrument, by compressing or removing clouds in images (Thompson et al. 2014). Similarly, cloud avoidance work has been undertaken on TANSO-FTS-2, using intelligent targeting to minimize observations effected by clouds (Suto et al. 2021). Greedy and graph-search-based algorithms are employed by Hasnain et al. to select the clearest sections of the sky for imaging (Hasnain et al. 2021).

The Smart Ice Cloud Sensing (SMICES) mission concept proposes using machine learning and artificial intelligence methods to target storms and clouds of scientific interest (Ogut et al. 2022a; Swope et al. 2021). Sponsored by the NASA Earth Science Technology Office (ESTO) under the Instrument Incubator Program 19 (IIP-19), SMICES, a combined radar/radiometer instrument, is designed to measure cloud ice and water vapor in the upper troposphere and lower stratosphere. It is able to operate at a very high rate, targeting images on the order of seconds, and utilizes multiple cloud labels to identify targets of scientific interest. The SMICES concept is currently in development for demonstration on an aircraft, but the ultimate goal is to deploy the algorithms on a small satellite in low-Earth orbit.

Recent work has demonstrated flight of advanced machine learning models on board small satellite platforms such as ESA's Φ-Sat (Giuffrida et al. 2022) and ESA's OPS-SAT (Labrèche et al. 2022). There exist several methods for satellite observation scheduling that use machine learning concepts. Chen et al. present an approach to scheduling agile satellite tasks that uses a recurrent neural network (Chen et al. 2019). Wei et al. develop a multi-objective method based on deep reinforcement learning and parameter transfer which considers both the failure rate and the timeliness when scheduling satellite tasks (Wei et al. 2021). These methods both assume that the satellite has only one imaging sensor (no lookahead), and neither method considers power consumption.

### 2.2 Dynamic Targeting

The purpose of dynamic targeting methods is to maximize the number of high value targets sampled by an Earth-observing satellite instrument, while complying with energy constraints. Prior algorithms for dynamic targeting were developed for use in both storm tracking and cloud avoidance scenarios (Candela et al. 2023). In the case of storm tracking, particular cloud types are more informative and scientifically valuable to sample, so those cloud types are given a higher priority. In other cases, high value scientific targets are only visible when there is no cloud cover, so sampling clear skies is given a higher priority.

Dynamic targeting methods use a model of a satellite with a primary sensor and a lookahead sensor that allows the satellite to see a limited view of its future path (see Figure 1). This configuration is seen in satellites such as the Smart Ice Cloud Sensing (SMICES) small-sat, which carries a radar to take targeted samples of deep convective ice storms, and a radiometer to find these storms along the path of the satellite (Ogut et al. 2022b; Bosch-Lluis et al. 2022).

Many existing methods for dynamic targeting follow a "greedy" policy, meaning they sample the location with the highest immediate reward, given some constraints. We will compare our methods to the following existing dynamic targeting approaches: random, greedy nadir, greedy lateral, greedy radar, greedy window, and a dynamic programming (DP) approach (Candela et al. 2023). See Figure 2 for an illustration of these approaches.

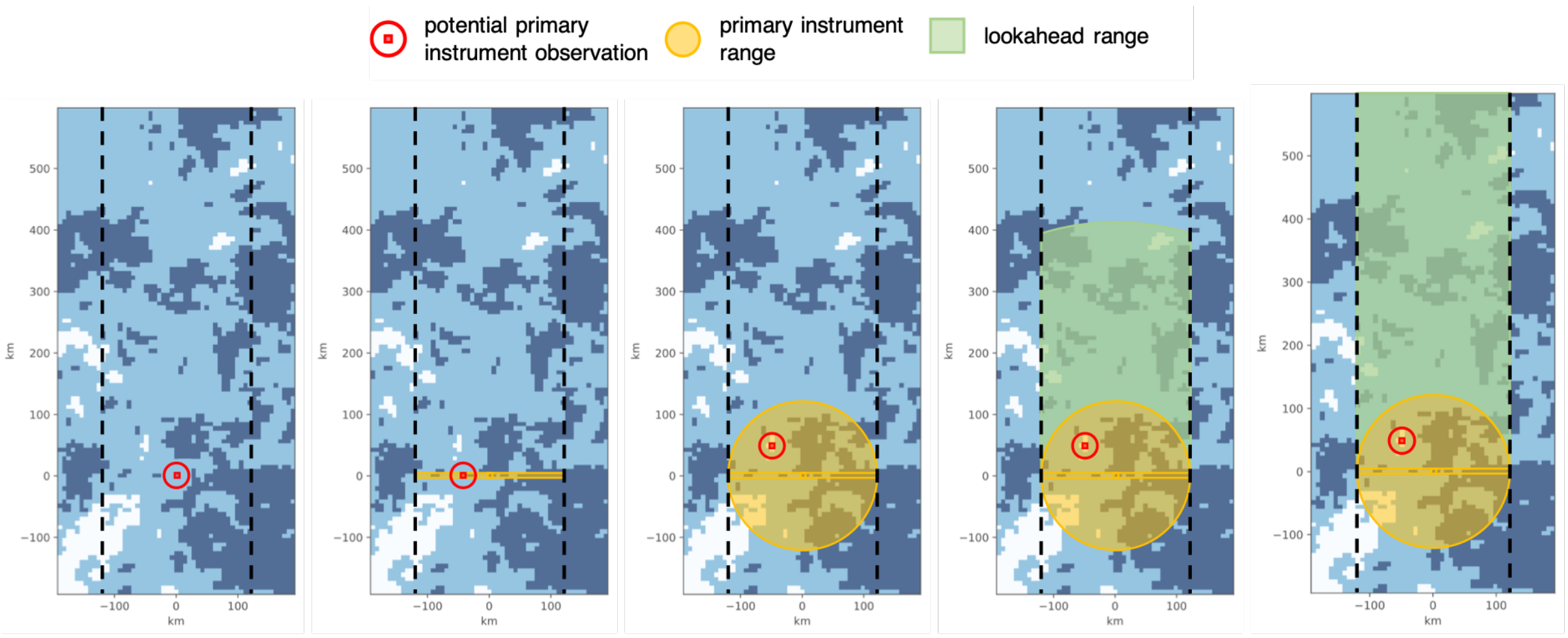

(a) random/greedy nadir (b) greedy lateral (c) greedy radar (d) greedy window (e) dynamic programming

*Figure 2: Existing dynamic targeting algorithms for intelligent satellite observations. Each has different knowledge of the environment. Our learning approaches have the same knowledge and sampling capabilities as greedy window (d). The dynamic programming approach (e) has full knowledge of future states and thus is used as a benchmark.*

Each approach has a different sampling space and knowledge of the environment. The greedy nadir sampling method always points the instrument directly downwards (at nadir) but decides whether to take a sample based on the state of charge (battery percentage) of the satellite and the target beneath the sensor. A random method is used as a lower bound, which also exclusively points the instrument at nadir, but randomly decides when to take a sample. The greedy lateral approach searches along the lateral axis of the sensor for the best target to sample and then determines whether to sample based on the best target type and the state of charge. The greedy radar method functions the same as the greedy lateral approach, but instead searches the entire reachable area of the primary (radar) sensor. The greedy window method uses the knowledge of the target types in the reachable sensor area and the lookahead sensor along with the state of charge of the satellite to determine whether to sample and where to point the primary sensor. Our methods have the same knowledge of the environment as the greedy window method (primary and lookahead sensor).

The dynamic programming approach uses the entire path of the satellite (not just what is visible to the sensors) to determine when and where to sample. The DP approach is used to generate an upper bound for performance comparisons but is not practical to implement on a satellite because the algorithm assumes that it has full knowledge of all future states of the world, which is not the case with real-world limited lookahead sensors.

### 2.3 Satellite Simulation

This work leverages an existing framework to model an Earth observation satellite with a primary radar sensor and a lookahead sensor (Candela et al. 2023). The simulation framework allows the user to specify the range and resolution of the primary and lookahead sensors, and the rate of charging/discharging of the satellite. For our simulation studies, we use a primary radar sensor with a range of 15º from nadir, and a lookahead sensor with a range of 45º from nadir. We use a resolution of 7 km/pixel for the lookahead images. Taking a sample discharges the satellite's battery by 5% of a complete charge, and it can recharge its battery by 1% at each time step (so effectively the satellite power discharges by 4% when taking a sample). This charge/discharge rate causes the satellite to keep the primary sensor off on average 80% of the time, and only take samples 20% of the time. For these studies, we assume that the lookahead sensor is always on and does not interfere with the charging rate of the satellite.

Our simulation is useful both for testing cloud avoidance to better observe the Earth's surface, and for testing storm hunting, or intelligently sampling cloud cover. The simulation calculates a feasible orbit for the satellite, and images are extracted from real historical data that represent local information that would be seen by the lookahead sensor given the time and the satellite's location. For cloud avoidance, we use data from the Moderate Resolution Imaging Spectroradiometer (MODIS) (Justice et al. 1998). Each pixel in the real satellite image is categorized into three classes: clear (high scientific reward), mid-cloud (medium scientific reward), and cloud (low scientific reward). For storm hunting, we use data from the Global Precipitation Measurement (GPM) mission, and classify the data into three storm types: no storm (low scientific reward), rainy anvil (medium scientific reward), and convective core (high scientific reward) (Hou et al. 2014; Integrated Multi-satellitE Retrievals for GPM (IMERG)).

# 3 METHODS

## 3.1 Problem Formulation

We model the dynamic targeting problem as a Markov Decision Process (MDP) as follows. A state $s \in S$ consists of the information that can be seen in the primary and lookahead sensors, and the current state of charge ($SOC \in [0\%, 100\%]$) of the satellite. An action $a \in A$ involves taking a sample with the primary instrument at a particular location, which receives some scientific reward $R \in \mathbb{R}$. The satellite transitions into the next state deterministically ($P(s, s') = 1$) by moving one discrete timestep $t$ forward along its orbit. However, the next state $s'$ cannot be fully known a priori as there is some new information that is yet to be observed by the lookahead sensor. The policy function $\pi$ is a mapping from state space ($S$) to action space ($A$).

We explore two different learning approaches that leverage ideas from dynamic programming to solve the dynamic targeting problem. These methods are reinforcement learning and imitation learning.

## 3.2 Dynamic Programming

Prior work uses a dynamic programming (DP) algorithm that produces optimal policies given a particular set of assumptions (Candela et al. 2023). However, DP is generally not deployable on missions because it is computationally expensive, it uses a lookahead sensor range that is physically unrealistic, and it requires information of future states.

We use dynamic programming as an “oracle” to generate an upper bound on performance for our models, but we will also use the concept of DP in our reinforcement learning and imitation learning methods. Algorithm 1 summarizes the DP formulation we will use throughout this work. This method has full knowledge of all images along the entire path that will be traversed. We use the standard memoization strategy, which works backwards from the last state in the environment (Bellman 2003). As the algorithm progresses, it “visits” every state in the environment (every image paired with every state of charge) and records the current plus future reward value for taking any action at that state. These reward values are stored in an array ($D$). After the algorithm terminates, to determine the optimal sequence of actions to take we simply look up the current state of the environment (based on the time step and the state of charge) and take the action that results in the highest future reward value according to $D$.

## 3.3 Reinforcement Learning

Reinforcement learning methods involve an agent that learns policies by interacting with the environment and receiving feedback via a reward function that incentivizes “good” actions and

**Algorithm 1: Dynamic Programming (Memoization)**

```
D ← Zeros(N_T, N_SOC, N_A)
for each t ∈ {T...1} do
    for each SOC ∈ {0...100} do
        for each A ∈ Actions do
            S ← Environment(t, SOC)        ▷ get state
            Execute A
            Observe R, S'        ▷ get reward and next state
            Observe SOC'        ▷ get new state of charge
            t' ← t + 1
            D(t, SOC, A) ← R + max_a D(t', SOC', a)
        end for
    end for
end for
```

*Algorithm 1: The dynamic programming (memoization) method considers every possible state of the environment to optimize reward.*

penalizes "bad" actions. Reinforcement learning is an iterative approach that improves gradually by learning from its mistakes and successes throughout many trials.

Here we use a variation of Q-learning to solve the dynamic targeting problem. Q-learning is a model-free reinforcement learning algorithm; it uses a learned lookup table called a Q-table which tells the agent which actions have the highest expected reward given a particular state of the environment. The Q-table contains a row for each possible state, and a column for every possible action. To follow the Q-learning policy ($\pi^\star$), we simply look up which action has the highest value in the Q-table for the current state of the environment, and execute that action. The algorithm "learns" using the update function:

$$Q(s_t, a_t) \leftarrow \quad Q(s_t, a_t) + \alpha \left( r_{t+1} + \gamma \max_a Q(S_{t+1}, a) - Q(s_t, a_t) \right)$$

where $Q$ represents the Q-table, $s_t$ is the current state of the world, $a_t$ is the last action taken, $r_{t+1}$ is the reward received from taking action $a_t$ in state $s_t$, $s_{t+1}$ is the state after taking action $a_t$, $\alpha$ is the learning rate, and $\gamma$ is a discount factor which tells the algorithm how heavily it should weight future rewards (Sutton and Barto 2020). Based on a parameter search, we choose $\alpha = 0.4$ and $\gamma = 0.99$. Using this update function, as the Q-learning algorithm experiences more states, it better learns which action leads to the highest reward from that state.

Q-learning is very sensitive to the total number of possible states, as this defines the size of the Q-table. If we define the environment such that it has fewer states, there is a higher likelihood that all states will be visited throughout training, which improves performance. As such, we cannot use a classified cloud image as an input to Q-learning because this method leads to too many state possibilities. Instead, we formulate a state vector to pick out the important features in our scenario. Those seven features are the state of charge (battery percentage) of the satellite, and six binary variables which represent the presence of the three cloud types in the primary sensor range, and the presence of the three cloud types in the lookahead sensor range. Thus, the Q-learning state representations for each scenario are:

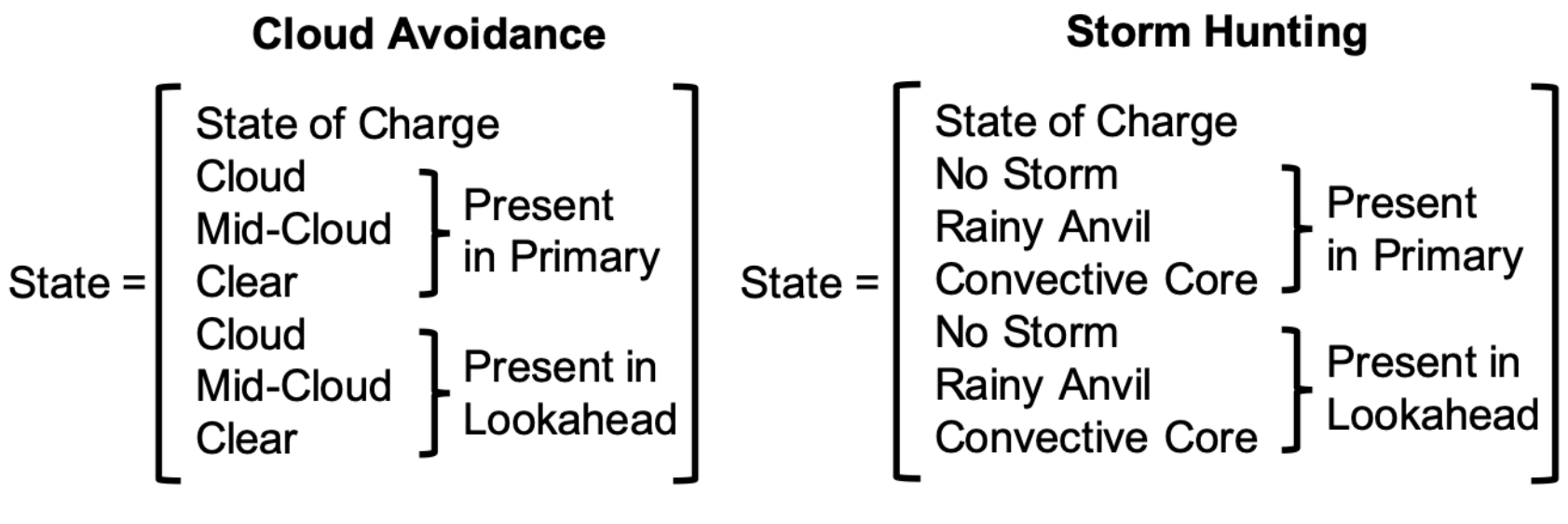

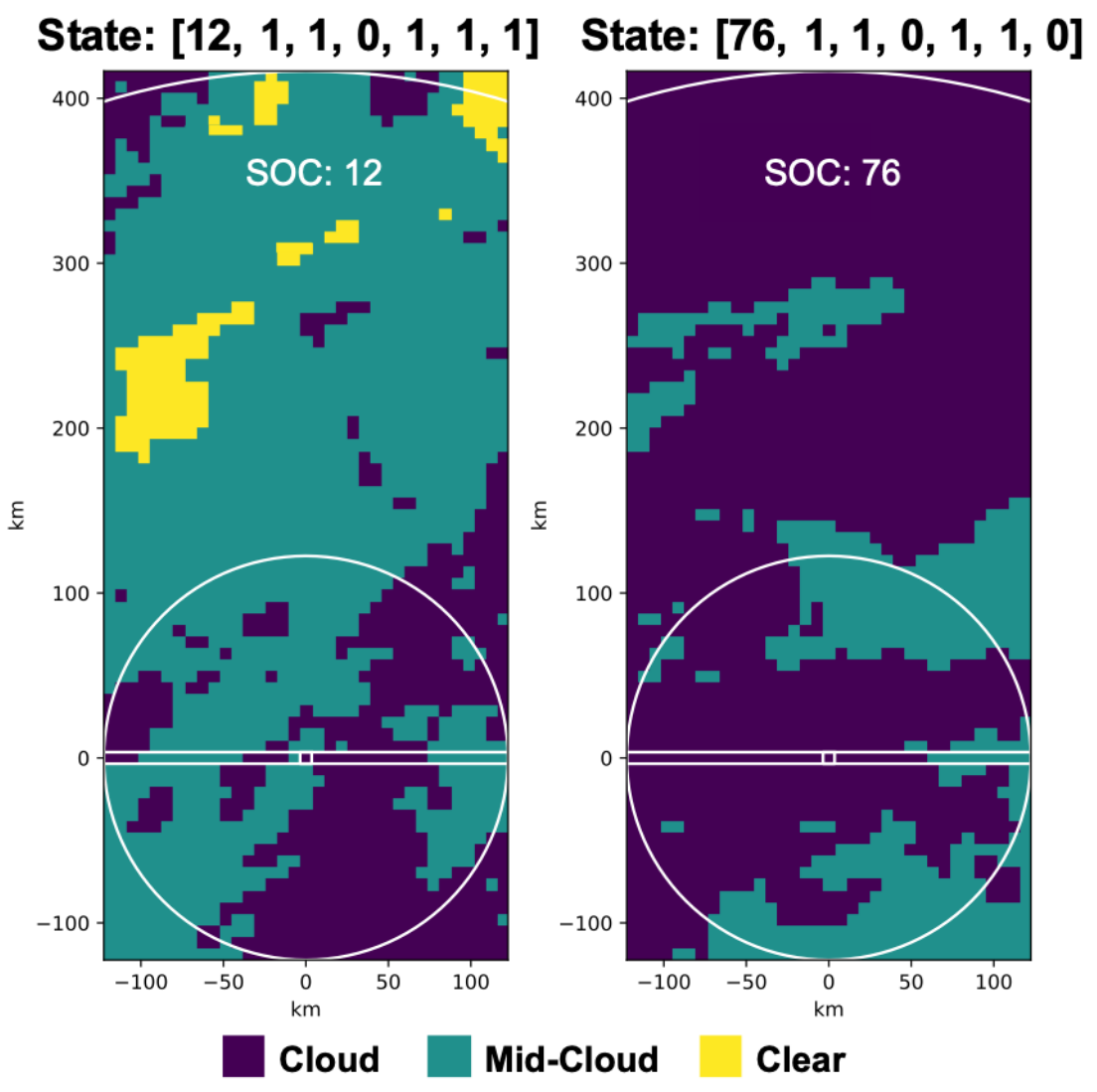

*Figure 3: Example of lookahead sensor images (after being classified into cloud types) and their corresponding Q-learning states for the cloud avoidance scenario. The white circle represents the range of the primary (radar) sensor.*

An example of cloud images and their corresponding Q-learning state representations are shown in Figure 3.

Our action for Q-learning is binary: to sample or not to sample. We do not choose *where* to sample using the Q-learning algorithm, just *when* to sample. Our formulation of the dynamic targeting problem does not consider the power draw or time taken to move the primary sensor. Thus, it is straightforward to determine where the satellite should sample, once we have determined at what states it should sample. If we are taking the "sampling" action, we simply measure the highest reward cloud type closest to nadir. Future work could address the problem of determining where to sample if we consider more real-world factors such as the time and energy necessary to move the sensor, and diminishing returns of sampling the same location.

Our formulation of the state of the environment for Q-learning results in $101 \times 2^6 = 6{,}464$ possible environment states. This means for the trained model to be fully generalizable, it must visit all 6,464 states during training. In general, Q-learning proceeds forward in time, taking actions and observing the environmental response, and updating its belief about the environment based on the rewards received. This formulation requires the Q-learning algorithm to naturally come across most or all of the possible states in its exploration, which can be unlikely or even impossible with a high number of states or limited data. Some Q-learning methods, like the $\epsilon$-greedy approach (see Algorithm 2), attempt to rectify this problem by sometimes choosing random actions to execute, therefore placing the environment in a new unseen state (Sutton and Barto 2020). In practice, we found that training the Q-table using the $\epsilon$-greedy method did not produce sufficient results; this method was not able to outperform the greedy window method from prior work, because it did not visit enough unique states to generalize well when encountering new testing data.

In order to remedy this issue, we use a strategy from dynamic programming (see Algorithm 1). The benefit of dynamic programming is that it visits every possible state of the environment, given particular training data. To improve the performance of our method, we use this feature of dynamic programming in our Q-learning algorithm. We begin iterating from the last timestep in the training data, and step backwards in time while varying the state of charge from 0 to 100, and observing the reward from taking each action from each of these states (see Algorithm 3). Thus, the Q-learning algorithm "experiences" taking every action from every

Algorithm 2: Q-Learning (ϵ-greedy)

```
Q ← Zeros(N_S, N_A)
for each Episode do
    S ← S_0                                  ▷ initial state
    p ← Random(0...1)
    if p ≤ ϵ then
        A ← Random(Actions)                  ▷ random action
    else
        A ← arg max_a(Q(S, a))               ▷ greedy action
    end if
    Execute A
    Observe R, S'                            ▷ get reward and next state
    Update Q                                 ▷ per equation 1
    S ← S'
end for
```

Algorithm 3: Q-Learning (DP-inspired)

```
for each t ∈ {T...1} do
    for each SOC ∈ {0...100} do
        for each A ∈ Actions do
            S ← Environment(t, SOC)          ▷ get state
            Execute A
            Observe R, S'                    ▷ get reward and next state
            Update Q                         ▷ per equation 1
        end for
    end for
end for
```

*Algorithms 2 and 3: Q-learning methods use a lookup table to determine the highest expected reward from taking a particular action.*

cloud image at all different levels of charge. This greatly increases the number of states that the Q-learning algorithm encounters while still using the same amount of satellite data for training. It is important to note that we are able to use this strategy because we have a simulation that allows us to step backwards in time through saved data. This strategy is not applicable to situations in which Q-learning is performed on a real system and not stored data. Although we are using the same framework as the dynamic programming approach, note that by learning the Q-table, we are able to use this method in general on unseen situations, whereas the dynamic programming approach will only provide a list of optimal actions for a specific sequence of data.

### 3.4 Imitation Learning

Imitation learning, also known as apprenticeship learning or learning from demonstration, consists of learning to replicate the actions of an expert agent. In this work the DP algorithm serves as the expert. Optimal policies can be approximated offline via learning, allowing for onboard inference, planning, and execution with more realistic mission resources.

We employ behavioral cloning, a form of imitation learning that learns a policy by using supervised learning on a set of expert demonstrations that is collected beforehand. The process is described in Algorithm 4 and consists of two main steps. First, the algorithm collects a set of expert demonstrations $\mathcal{D}^\star$ that consists of state-action pairs: $\mathcal{D}^\star = \{(s_0, a_0^\star), (s_1, a_1^\star), \ldots\}$. Behavioral cloning methods typically collect non-interrupting sequences of states and actions (also known as a trajectories), but here we randomly discard some state-action pairs; this is for efficiency purposes as we later show how relatively small datasets can yield good results. Finally, the method trains a supervised learning model using the expert demonstrations in order to learn a policy $\hat{\pi}$ that approximates the expert policy $\pi^\star$.

We use a neural network for supervised learning, specifically a multilayer perceptron (MLP) with an architecture that is specified in Table 1.

| Layer | Units | Activation Function |
|---|---|---|
| input | 13 | N/A |
| hidden 1 | 32 | rectified linear unit |
| hidden 2 | 16 | rectified linear unit |
| hidden 3 | 8 | rectified linear unit |
| hidden 4 | 4 | rectified linear unit |
| output | 1 | sigmoid |

*Table 1: Architecture of the multilayer perceptron network used for behavioral cloning.*

Algorithm 4: Behavioral Cloning (DP expert)

$\mathcal{D}^\star \leftarrow \{\}$ ▷ empty dataset
**for** each $t \in \{1...T\}$ **do**
  **for** each $SOC \in \{0...100\}$ **do**
    $p \leftarrow Random(0...1)$
    **if** $p \leq \epsilon$ **then** ▷ random selection
      $S \leftarrow Environment(t, SOC)$ ▷ get state
      $A^\star \leftarrow \arg\max_a D(t, SOC, a)$ ▷ get action
      $\mathcal{D}^\star \leftarrow \mathcal{D}^\star \cup (S, A^\star)$ ▷ save state-action pair
    **end if**
  **end for**
**end for**
$\hat{\pi}_\theta \leftarrow NeuralNetwork.Train(\mathcal{D}^\star)$ ▷ learn policy

*Algorithm 4: Behavioral cloning is a method of imitation learning which learns to replicate the actions of an expert.*

The goal of the neural network is to minimize the difference between the actions drawn from $\hat{\pi}$ and the corresponding demonstrations from $\pi^\star$:

$$\hat{\theta} = \underset{\theta}{\operatorname{argmin}} \sum_{(s,a^\star)\in\mathcal{D}^\star} \mathcal{L}\left(\hat{\pi}_\theta(s), a^\star\right),$$

where $\mathcal{L}$ is a loss function (e.g., mean squared error), $\hat{\pi}_\theta(x)$ is the action drawn from the learned policy for state $s$, $a^\star$ is the expert's action for the same state, $\theta$ is the set of parameters (in this case neural network weights) that define the learned policy $\hat{\pi}_\theta$, and $\hat{\theta}$ is the set of learned parameters after solving the optimization problem.

For this approach, states $S$, actions $A$, and the policy function $\pi$ are formulated with some differences because behavioral cloning does not face the same limitations as Q-learning, that is, the need for discrete states and actions in conjunction with a manageable Q-table size. States are feature vectors that contain more information than in the Q-learning case and are normalized between 0 and 1:

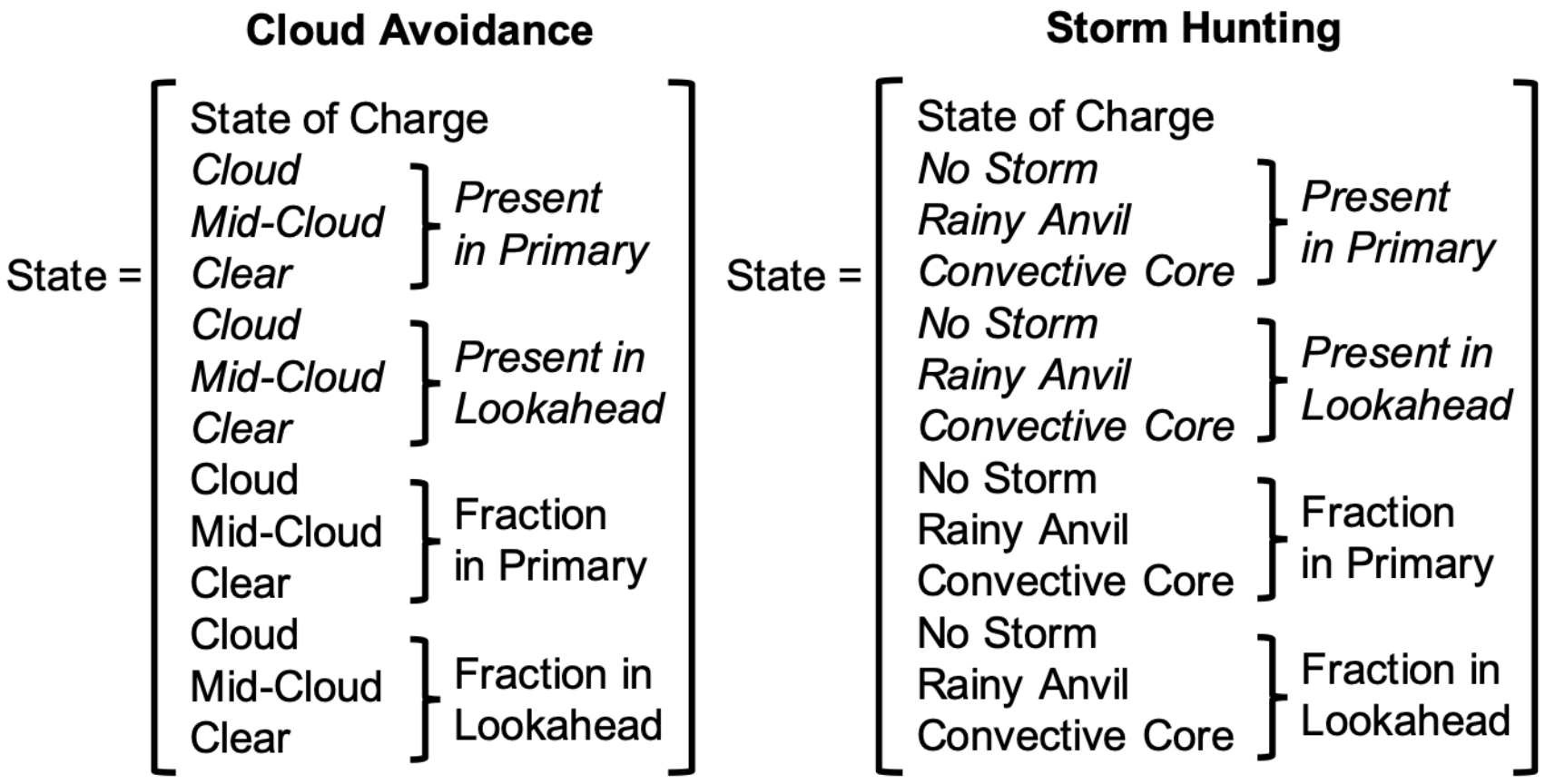

Actions are now stochastic as they are randomly drawn from the policy that was learned by the neural network. That is, $a_t \sim \hat{\pi}_\theta(s_t)$, where $\hat{\pi}_\theta(s_t) \in [0,1]$ is the probability of triggering an observation $a_t \in \{0,1\}$ given a state $s_t \subset \mathbb{R}^{13}$.

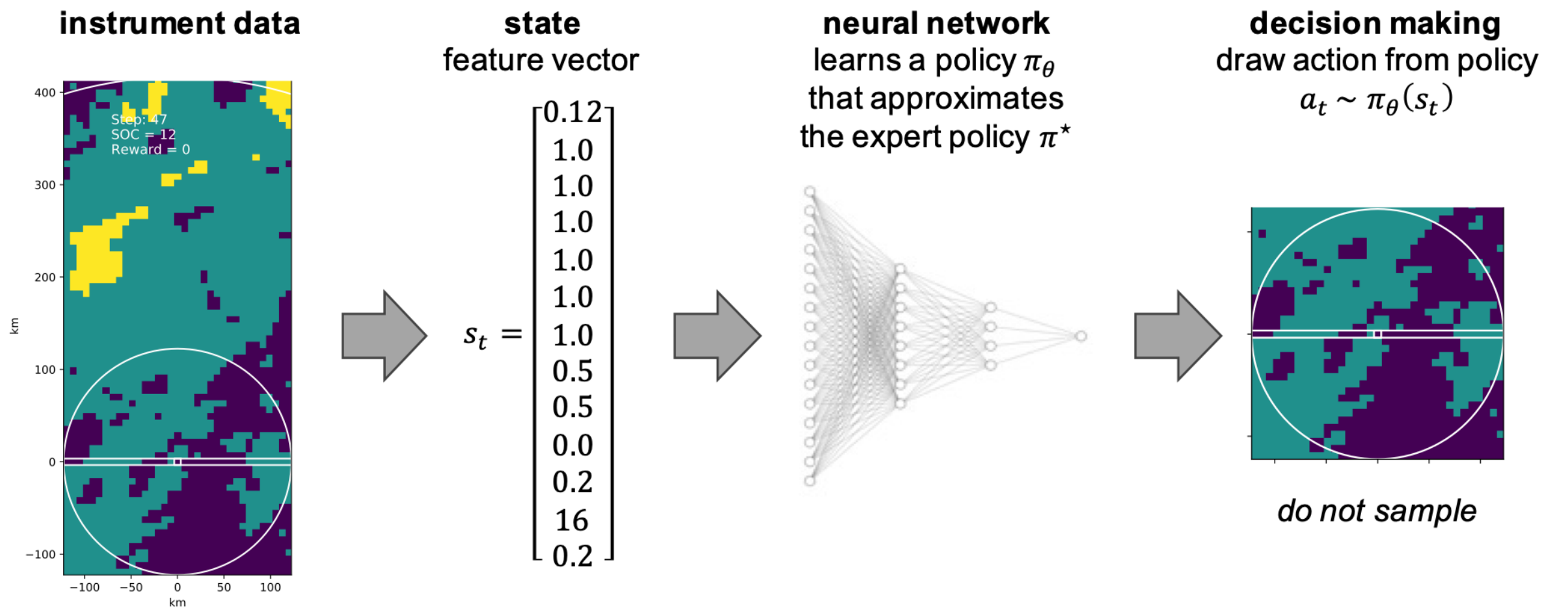

*Figure 4: Dynamic targeting via behavioral cloning inference. First, instrument data is processed to generate a state vector. Then, a neural network is used to decide when to trigger observations based on what it previously learned from expert demonstrations. In this example the network learns that it is best to wait and save resources for more valuable future observations.*

# 4 EXPERIMENTS

We test our methods in two scenarios: cloud avoidance and storm hunting. We use 39 datasets from the MODIS mission for cloud avoidance and 10 datasets from the GPM mission for storm hunting. Each dataset was collected at a different week of the year and represents one day worth of satellite images (86,400 images per dataset). We use half of the datasets for training our two learning models and the other half for testing. The rewards for each scenario are as follows:

- Cloud Avoidance: off (no sample) = 0, cloud = 1, mid-cloud = 10, clear = 100
- Storm Hunting: off (no sample) = 0, no storm = 1, rainy anvil = 10, convective core = 100

We first experiment with using different amounts of data to train our learning models, and ultimately choose the one with the best and most consistent performance across the testing data. We then compare these learning methods to prior work that consists of different heuristic algorithms.

# 5 RESULTS

## 5.1 Training

We found that in most cases, using a full dataset (86,400 images) to train each Q-learning model resulted in a higher or equal total reward on the testing data than using a fraction of the data from each dataset. In short, using more data to train improved performance. Figure 5 illustrates the effect of the amount of training data on the reward value achieved by each set of models. The "percent of total reward" is determined in comparison to the dynamic programming approach, which gives the optimal sequence of actions and therefore the optimal reward. We see that the average performance of our models levels out at 20,000 images worth of training data in the cloud avoidance scenario, and at just 3,000 images worth of training data for the storm hunting scenario. In the cloud avoidance case, using all images to train our model outperforms the average reward value from the models trained on one full dataset (86,400 images), but actually falls short of the best-performing models trained on 5,000 to 86,400 images. In the storm hunting scenario our performance actually decreases when we use all the training images, probably due to overfitting to the training data in this case. Overall, these experiments indicate that our Q-learning method can be trained effectively on a small amount of data (just 3,000 to 20,000 satellite images), though the amount of image data needed depends on the scenario tested.

We observed similar results for the behavioral learning approach (Figure 6). Performance increases as the neural network is trained with more data, in this case demonstrations from the DP expert. However, performance converges without too many examples. Each full DP dataset consists of 86400 × 101 = 8726400 different possible states (images and SOC). The training sets that we use in this experiment represent a small fraction of this total, ranging from 0.34% to 2.41%. An important consideration is that while training data is randomly sampled, in practice we also balance it in such a way that the three different cloud types are seen and represented more or less equally. This is to reduce learning bias of the neural network, and is one possible explanation for such a quick convergence in performance.

In both cases, we found that some datasets in our training data contained more unique states than others, and thus led to a consistently higher total reward on all testing data as compared to models trained on datasets with fewer unique states.

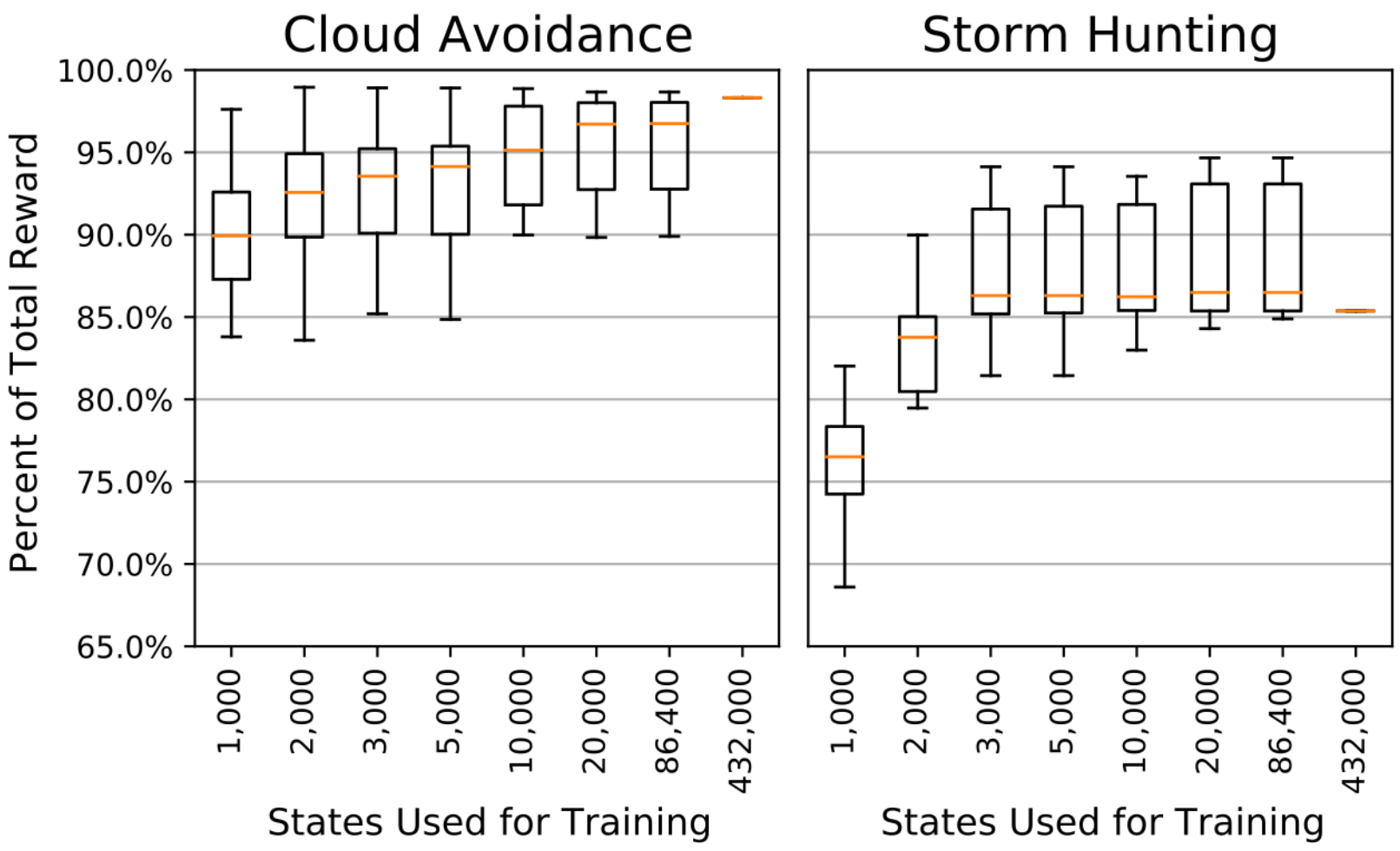

*Figure 5: Percent of total possible reward achieved by each set of Q-learning models trained on different amounts of data. Each set of models is trained on the first N states in each training dataset. A full dataset contains 86,400 images.*

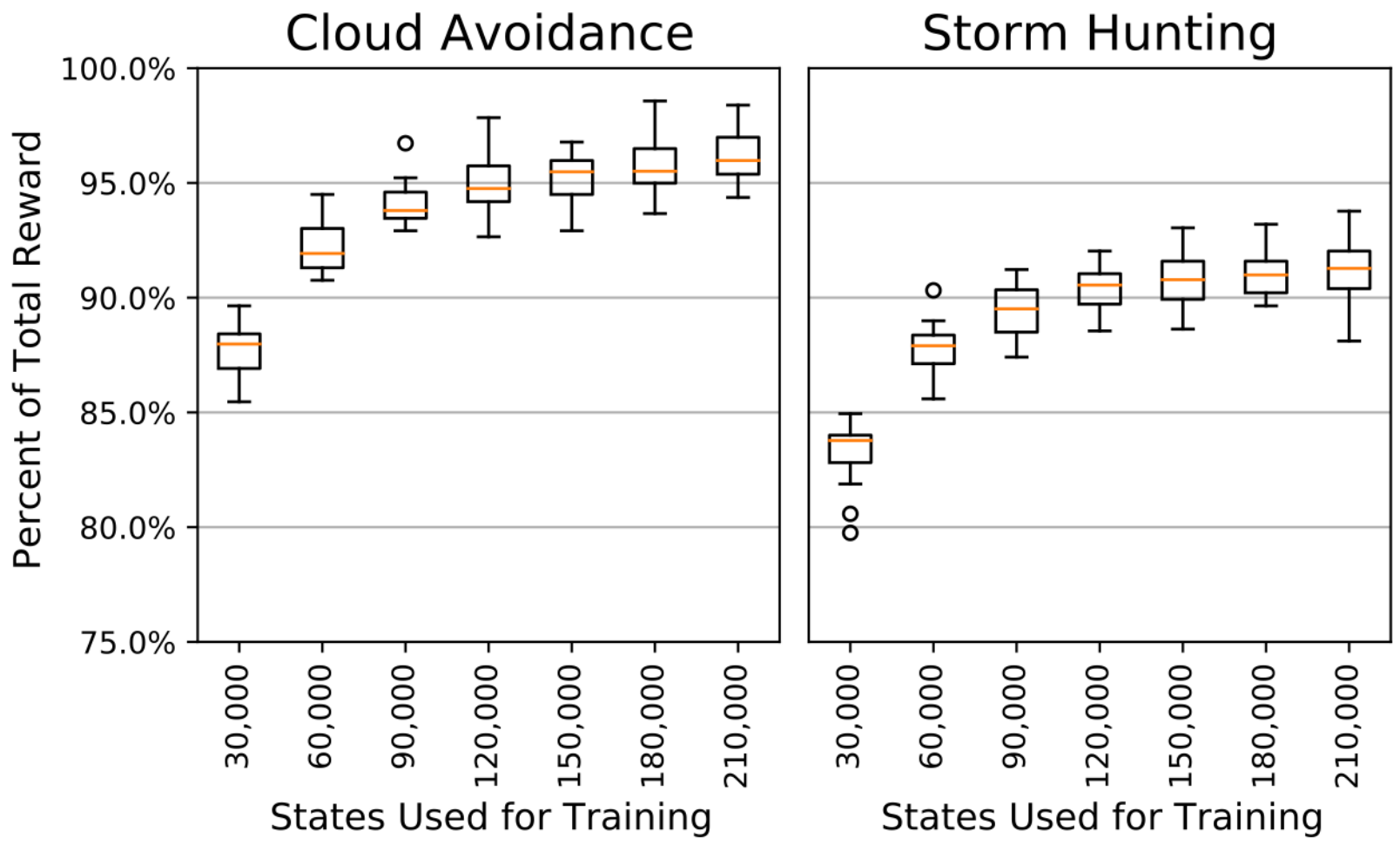

*Figure 6: Percent of total possible reward achieved by the behavioral learning neural network when trained on different amounts of randomly sampled data.*

### 5.2 Algorithm Comparison

We compare our two learning approaches to existing planning methods designed to solve the dynamic targeting problem: random, greedy nadir, greedy lateral, greedy radar, greedy window, and dynamic programming as an “oracle,” i.e. an upper bound on performance (for a description of these methods, see Section 2). Table 2 shows how often each planning algorithm sampled each cloud type over all testing data. Note that for both scenarios, the cloud types that have a high scientific reward value are generally rarer to encounter, while cloud types with a lower scientific reward are much more common. In addition, due to the energy constraints of the satellite, we expect that each algorithm will have to spend 80% of the time “off” or not sampling, to recharge the battery.

We see that in all cases, our Q-learning and behavioral cloning approaches sample high reward cloud types more frequently than any other planning method. Only the dynamic programming algorithm outperforms them in this regard, but the DP method is unrealistic to use in a real mission scenario and simply provides an upper bound on performance. The learning methods sample fewer low and mid-reward clouds than the other methods. This is a consequence of two factors: 1) sampling more high-reward cloud types means the learning methods have less time to sample low-reward cloud types, and 2) the reward function used by the learning algorithms encourages sampling higher reward cloud types ten times more than the next lowest reward cloud type. In other words, our methods are trained to maximize reward rather than to diversify samples, so this result is expected. Interestingly, we can see from Table 2 that the Q-learning method never samples Cloud (Reward = 1) types in the cloud avoidance scenario, and spends about 84% of its time not sampling, versus the expected 80% for recharging. This behavior is most likely due to the density of high reward cloud types in the training data; the Q-learning method generally learns to conserve power when it can only see low-reward cloud types, since based on the training data, there is likely to be a higher reward cloud type encountered soon.

Out of all the methods tested, Q-learning best utilizes the data in the lookahead sensor to take the most scientifically rewarding samples. Behavioral cloning is the second best. This difference is probably due to the fact that Q-learning uses information from future state rewards whereas behavioral cloning only relies on current state information. Hyperparameter tuning might be another explaining factor.

| **Cloud Avoidance** | **Random** | **Greedy Nadir** | **Greedy Lateral** | **Greedy Radar** | **Greedy Window** | **Behavioral Cloning** | **Q-Learning** | **Oracle (DP)** |
|---|---|---|---|---|---|---|---|---|
| Off (Reward = 0) | 80.37% | 79.98% | 79.98% | 79.98% | 80.00% | 79.98 % | 84.42% | 80.00% |
| Cloud (Reward = 1) | 13.31% | 10.07% | 6.43% | 4.28% | 4.29% | 4.16 % | 0.00 % | 3.67% |
| Mid-Cloud (Reward = 10) | 4.33% | 7.37% | 9.09% | 9.35% | 7.93% | 7.34 % | 6.52% | 7.24% |
| Clear (Reward = 100) | 1.99% | 2.57% | 4.50% | 6.39% | 7.78% | 8.52 % | 9.06% | 9.08% |
| **Storm Hunting** | | | | | | | | |
| Off (Reward = 0) | 80.37% | 79.99% | 79.99% | 79.99% | 80.02% | 79.99 % | 80.17% | 80.00% |
| No Storm (Reward = 1) | 18.82% | 18.01% | 15.47% | 13.17% | 13.17% | 13.14 % | 13.12% | 12.28% |
| Rainy Anvil (Reward = 10) | 0.78% | 1.88% | 4.23% | 6.23% | 5.79% | 5.71 % | 5.41% | 6.40% |
| Convective Core (Reward = 100) | 0.03% | 0.12% | 0.31% | 0.60% | 1.01% | 1.16 % | 1.30% | 1.32% |

*Table 2: Percentage of time spent sampling each cloud type by each planning method in both scenarios.*

| | Percent of Total Possible Reward (Cloud Avoidance) | Percent of Total Possible Reward (Storm Hunting) |
|---|---|---|
| **Random** | 25.91% | 14.12% |
| **Greedy Nadir** | 34.57% | 23.29% |
| **Greedy Lateral** | 55.58% | 42.45% |
| **Greedy Radar** | 74.86% | 65.18% |
| **Greedy Window** | 87.50% | 82.57% |
| **Behavioral Cloning** | 95.84% | 91.27% |
| **Q-Learning** | 98.67% | 94.66% |
| **Oracle (DP)** | 100.00% | 100.00% |

*Table 3: Average percent of total possible reward achieved by each planning method.*

We also examine the total reward achieved by each planning method. Table 3 shows the reward attained by each algorithm as a percent of the total possible reward. This total possible reward is the reward achieved by dynamic programming, which provides the optimal sequence of actions for each testing dataset. In the cloud avoidance scenario, Q-learning and behavioral cloning attain 98.67% and 95.87% of the possible reward on average (respectively), whereas the best dynamic targeting method from prior work (greedy window) achieves only 87.50% of the possible reward on average. In the storm hunting scenario, Q-learning and behavioral cloning achieve 94.66% and 91.27% of the possible reward where the greedy window method achieves 82.57%.

Every method takes on the order of microseconds to plan a sample (with the exception of dynamic programming, which is not feasible to use on a real system), so they will each comfortably run in real time on a satellite that is only capable of sampling once per second. Small differences in runtime between methods are thus not relevant.

# 6 CONCLUSIONS AND FUTURE WORK

This work presents two learning-based planning methods for dynamic targeting to improve the science return of Earth-observing satellites. These two methods build on dynamic programming and consist of reinforcement learning (Q-learning) and imitation learning (behavioral cloning), respectively. Simulation results demonstrate that both learning-based approaches perform better than existing heuristic methods, and also close to optimal. Additionally, both learning methods can be effectively trained with relatively small amounts of data.

Future work will consider more realistic satellite factors and instrument constraints, as well more interesting reward models. Additionally, we would like to use full images as inputs rather than manually-engineered state vectors, thus preserving information that can potentially lead to better decisions. We would also like to use reinforcement learning methods such as deep Q-networks (DQN) and proximal policy optimization (PPO) that easily allow for continuous state representations. Finally, we plan to deploy and test these algorithms on different satellite platforms, especially those that have flight processors that support deep learning.

# 7 ACKNOWLEDGEMENTS

The research was carried out in part at the Jet Propulsion Laboratory, California Institute of Technology, under a contract with the National Aeronautics and Space Administration (NASA) (80NM0018D0004). This work was supported by the Earth Science and Technology Office (ESTO), NASA, and by NASA's Lunar Surface and Instrumentation and Technology Payload (LSITP) program (award 80MSFC20C0008 to Astrobotic Technologies).